# A Scalable and High Availability Solution for Recommending Resolutions to Problem Tickets

Harish Saragadam
Global AI Accelerator, Ericsson
Bangalore, India
harish.s@ericsson.com

Chetana K Nayak
Global AI Accelerator, Ericsson
Bangalore, India
chetana.k.nayak@ericsson.com

Joy Bose
Global AI Accelerator, Ericsson
Bangalore, India
joy.bose@ericsson.com

*Abstract*— Resolution of incidents or problem tickets is a common theme in service industries in any sector, including billing and charging systems in telecom domain. Machine learning can help to identify patterns and suggest resolutions for the problem tickets, based on patterns in the historical data of the tickets. However, this process may be complicated due to a variety of phenomena such as data drift and issues such as missing data, lack of data pertaining to resolutions of past incidents, too many similar sounding resolutions due to free text and similar sounding text. This paper proposes a robust ML-driven solution employing clustering, supervised learning, and advanced NLP models to tackle these challenges effectively. Building on previous work, we demonstrate clustering-based resolution identification, supervised classification with LDA, Siamese networks, and One-shot learning, Index embedding. Additionally, we present a real-time dashboard and a highly available Kubernetes-based production deployment. Our experiments with both the open-source Bitext customer-support dataset and proprietary telecom datasets demonstrate high prediction accuracy.

I. INTRODUCTION

The problem of recommending resolutions for problem tickets or incidents on the basis of historical data is an important problem for service users, including telecom operators. Typically, service desks have dedicated manual teams that perform triaging of the issues and root cause analysis, and recommending a solution can take several hours end to end. Using machine learning models to recommend resolutions can save significant time and manpower of the operators by recommending solutions based on historical incident data. However, real-world application involves addressing several practical challenges:

- Diverse ticketing formats across service desks.
- Missing resolution fields and inconsistent text input.
- Data drift and evolving incidents.
- Customer preference for dynamic resolution recommendations (e.g., chatbots).
- Requirements for high availability and detailed analytics.

The problem statement can be broken down as follows: We are given a dataset of historical data mapping problem ticket descriptions, with unique incident IDs, with resolutions to the problems, along with unique resolution IDs. Training a model on this dataset, for a new incoming problem ticket, recommend the most relevant resolution along with its resolution ID. Here, both the problem description and resolution are in the form of short text.

This problem can be solved using a variety of models and techniques utilizing Natural Language Processing (NLP). One such solution by Ferland et al. [1] explores a hybrid approach with unsupervised clustering and deep learning embedding. However, on discussion with various clients there are a few problems with developing a generic solution that can work for multiple customer systems:

- Different service desks have different types of fields and the solution may have to be retrained from scratch or may need to adapted to fit the specific needs of the service desk.
- Resolution and other fields needed for the existing solution are often not present in the service desk systems and the solution cannot work without them
- Sometimes, the operators manning the service desk are not trained uniformly and may input resolutions as free text, leading to slightly different wordings used for the same resolutions.
- The trained machine learning (ML) models may be sensitive to data drift. Older data pertaining to incidents whose root cause has already been fixed, may be redundant and hence not relevant any more. In particular, if new issues come that it has not seen during training, the model may give wrong resolutions.
- Sometimes, just providing the most relevant resolution based on history may not meet the needs of the service desk They may prefer to have a chatbot type of solution instead.
- The service desks may have other requirements such as needing to be high availability, and needing to have access in the form of a dashboard for customers to analyze trends and get more information about patterns in problem tickets: apart from just the resolutions.

The sample data has the following structure:

*Incident Number, Description, Submit Date, Resolution, Resolution ID*

An example of historical problem tickets data is as follows:

*INC001, Network not available in my area, 01-07-2024 1:16 AM, Network connectivity is now working fine, RES001*

*INC002, Recharge not updating in my account ,01-07-2024 12:23 PM, Billing system issue, Customer recharge updating properly now, RES002*

The previously mentioned problems are quite common for many service desk customers, yet the solution is often not trivial for the following reasons:

- Often the operations employees use free text for resolution, resulting in incidents that have different strings for the same resolution. That makes it difficult to train the ML model on such supervised data.
- The dataset may be unbalanced, with some problems occurring quite frequently and others being rare.
- Sometimes, the service desks may have lack of fields such as Resolution and Resolution Ids in the tickets data, making it difficult to train any supervised model.
- The current solution does not have high availability, which many customers prefer. There is a manual process of switching to the secondary backup by using a script.

In this paper, we seek to resolve the common problems using a variety of methods to overcome the mentioned issues. We try clustering to group similar sounding resolutions and map them to the same resolution IDs. In cases where data is missing or where there is no unique field or ID for resolution, we try generating unique numbers as resolution IDs. We also explore the use of large language models or LLMs, in the cases where the customer prefers a chatbot to resolve their issues rather than just a fixed resolution ID field. Additionally, we integrate a real-time analytics dashboard and deploy a highly available solution leveraging Kubernetes and Helm charts.

The rest of this paper is organized as follows: in section 2 we discuss related work in the domain of recommending resolutions for problem tickets. In section 3 we discuss the components of our proposed solution, focusing on the ML models in the ensemble. In section 4 we discuss the implementation details for various modules including the ML model and PowerBI dashboard. In section 5 we discuss preliminary results obtained with an open-source dataset and a proprietary dataset. In section 6 we discuss a few alternative approaches using newer ML models. Finally, Section 7 concludes the paper and details some directions for future work.

## II. Related work

The automation of ticket resolution and classification has seen significant research interest in recent years, particularly within telecommunications and IT support domains. One of the foundational challenges addressed is the high volume of trouble tickets and the manual effort required for triaging, classification, and resolution.

Marzo (2021) proposed a multi-stage text ranking approach using BERT-based models for retrieving solutions to trouble tickets in telecom systems. Her work focused on balancing accuracy and latency through an initial retrieval stage using Sentence-BERT and a subsequent re-ranking phase with monoBERT. This architecture significantly outperformed traditional retrieval models like BM25, demonstrating that transformer-based models are highly effective in industrial-scale troubleshooting environments.

Similarly, Ferland et al. [1] explored a hybrid NLP model that combines unsupervised clustering and supervised deep learning embeddings to resolve trouble tickets. Their approach leverages ensemble methods to generate resolution recommendations based on historical ticket data. Notably, their system does not generate new text but retrieves previously validated resolutions, enhancing consistency and reliability in practical deployments. There's is the base model we use for this paper. Following a similar direction, we enhance their approach addressing our specific use-cases and challenges.

In the context of large-scale customer service systems, Apple researchers introduced TaDaa, a deep learning-based ticket assignment system that utilizes transformer architectures to suggest the appropriate group, resolver, and similar past tickets. Their results showed top-3 accuracy of over 95% for group suggestions and 79% top-5 accuracy for resolver assignments. This highlights the viability of deep learning and ANN-based similarity search for efficient ticket routing and triage.

Research by Rocha et al. focused on classifying customer complaints in broadband services using traditional ML classifiers. Their work emphasized the importance of tailored preprocessing techniques and demonstrated that the Extra Trees classifier outperformed others with an accuracy of 89%, underscoring the continuing relevance of classical supervised models for structured classification tasks.

Other studies, such as Almarzooqi (2025), proposed hybrid systems combining rule-based and machine learning models for prioritization and routing of IT tickets. These systems leverage TF-IDF vectorization, ensemble models, and real-time feedback loops to handle class imbalance and dynamic ticket categories, providing adaptable frameworks suitable for live ITSM environments.

Overall, while traditional machine learning models remain useful for simpler classification problems, the trend has shifted towards leveraging deep learning and transformer-based models for more complex ticket resolution tasks. These models offer improved semantic understanding, scalability, and adaptability, making them well-suited for large-scale deployment in service desk operations.

## III. Proposed Approach

This section details the machine learning models and feature engineering strategies used to address the ticket resolution prediction task. Our approach combines unsupervised and supervised learning methods with ensemble modeling to produce robust recommendations. A block diagram of the proposed approach is provided in Fig 1.

Our comprehensive solution architecture comprises:

- Resolution ID Generation: For missing or inconsistent data, clustering algorithms (K-Means, GMM) generate unique resolution IDs.
- Supervised Classification: Trained models (LDA, Siamese network, Index embedding) classify and predict resolutions.
- Inference and Confidence Assessment: Confidence thresholds identify new, potentially unseen incidents, and default resolutions provide a baseline fallback mechanism.

- Dashboard Integration: A real-time PowerBI dashboard provides metrics including average resolution times, clustering effectiveness, and algorithm performance.
- High Availability Deployment: Kubernetes orchestrates containerized applications, with Helm charts facilitating reproducible and automated deployments.

A brief description of the feature engineering steps and the stacked ensemble ML algorithm, largely based on [1] is provided below.

**Feature Engineering**

LDA (Latent Dirichlet Allocation): For topic modelling using LDA, we first apply *lemmatization* and *stemming* to normalize the vocabulary by reducing inflected or derived words to their root forms. This process improves generalization by allowing the model to learn from fewer data instances. We also apply frequency-based filtering to remove words that are either too rare (potentially noisy and uninformative) or too frequent (stopwords or overly generic terms). This enhances the discriminative power of the model.

Subsequently, we extract features using TF-IDF (Term Frequency–Inverse Document Frequency), along with part-of-speech (POS) tagging and n-grams (bigrams and trigrams). These features serve as the input representation for downstream machine learning models, enabling the capture of contextual and syntactic information from ticket descriptions.

FastText Embeddings (for LSTM and One-Shot Learning): For models based on neural embeddings, we use pretrained FastText vectors. FastText not only captures the semantic similarity between words through distributional representations but also handles out-of-vocabulary and misspelled words by leveraging subword information. This allows the model to assign embeddings to noisy inputs and improves generalization on real-world text data.

Index Embedding Models: For index-based neural architectures, we employ lightweight preprocessing including tokenization, stemming, and lemmatization of both issue descriptions and resolutions. These normalized tokens are then mapped to unique identifiers to facilitate efficient training of the embedding layers.

**Machine Learning Models:** A brief description of the ML Models is given below.

**Topic Modelling with LDA:**

Latent Dirichlet Allocation (LDA) is a probabilistic generative model that assumes each document (ticket) is composed of multiple latent topics, and each topic is characterized by a distribution over words. LDA operates under the bag-of-words assumption, making it well-suited for non-sequential and terse data such as ticket logs or system commands.

To train the LDA model, we perform the following steps:

1. Normalize the corpus using lemmatization and stemming.
2. Filter words based on document frequency thresholds.
3. Apply LDA to learn a fixed number of latent topics.
4. Assign the dominant topic to each ticket and rank associated resolutions based on frequency within that topic.
5. For new tickets, infer the topic distribution and recommend the most probable resolutions.

While LDA offers interpretable clusters, it is known to be sensitive to data drift. To address this, we retrain the model regularly and restrict training to recent time windows to maintain consistency with evolving ticket distributions.

**One-Shot Learning via Siamese Networks:**

Given the high diversity and imbalance in real-world ticket datasets, where many resolution classes have few labelled instances, we adopt a Siamese neural network for one-shot learning. Siamese networks consist of two identical sub-networks with shared weights that encode input pairs into a common feature space.

To train this model, we employ triplet loss, which encourages the distance between similar tickets (anchor and positive) to be minimized, while maximizing the distance from dissimilar ones (negative samples). Input embeddings are derived from LDA vectors and FastText features.

To mitigate the scarcity of training examples per class, we apply data augmentation by replacing a few words in existing tickets while preserving labels. This process boosts model generalization and enables effective learning of fine-grained distinctions. The final encoded representation is a 32-dimensional vector learned through a 6-layer perceptron, with the lower layers frozen during fine-tuning.

**Index Embedding Neural Networks:**

Index embedding models learn to represent each unique issue and resolution description as low-dimensional latent vectors. Unlike word-level models, the embedding here is computed for entire ticket texts and resolution texts.

The model architecture includes two embedding matrices: one for issue descriptions and one for resolution descriptions. Training is performed using both positive samples (known issue-resolution pairs) and negative samples (random mismatches). The loss function penalizes the model for assigning high similarity to incorrect pairs.

While this method scales well and provides efficient inference, it suffers from the cold-start problem—i.e., inability to predict resolutions for unseen issues. To mitigate this, similarity-based retrieval from historical data (using cosine similarity or nearest-neighbour search) is employed as a fallback mechanism.

**Ensemble Learning:**

To create the new training dataset, we took the recommended resolutions from each base model, including LDA, LSTM, One Shot, and index embedding, and then applied logistic regression on this new dataset to produce a final recommendation. The stacking approach we used allowed us to leverage the strengths of each individual model, while also mitigating their weaknesses.

By combining the recommendations from multiple models, we were able to produce a more accurate and robust recommendation system.

## IV. PROPOSED SOLUTION

The outline of our proposed solution consists of the following:

- In case of no resolution ID available, we generate our own resolution IDs based on clustering the tickets data
- We then train a supervised model to predict the resolution id
- In the case where the model has not seen the data during training, we identify it based on prediction confidence and show extra relevant information in the dashboard
- A dashboard, built using a tool such as PowerBI, that shows insights such as: Average time taken to resolve ticket, Most common text identification for the new tickets, Data such as similar tickets to the current ticket
- We have a fully highly available solution based on state of the art tools including Argo and Kubernetes.

The solution comprises of the following:

A high level overview of our approach is as follows:

- We use data from the existing systems to work on the solution
- We have divided the data into 80:20 split
- We use clustering to generate clusters on the training set and cluster number we use as proxy for the resolution ID
- We then use the existing ML classification to classify the clusters in the test set data
- If the confidence measure is <30% after summing up the top label probabilities we generate additional analytics: such as similar tickets

### A. Training phase

The training process begins by ingesting historical incident data, which includes ticket descriptions, timestamps, and—where available—corresponding resolution texts or resolution IDs. Given the often-missing or inconsistent resolution fields in real-world service desk systems, a crucial preprocessing step involves generating proxy resolution IDs. This is accomplished using unsupervised clustering techniques such as K-Means or Gaussian Mixture Models (GMM), applied to resolution text to group semantically similar resolutions. Each cluster is then treated as a surrogate resolution class.

The training pipeline is composed of the following stages:

**Data Splitting**: The dataset is divided into 80% training and 20% test data to ensure unbiased evaluation.

**Resolution ID Generation**: If explicit resolution IDs are absent, clustering is performed on the training data's resolution texts. Each resulting cluster is assigned a unique synthetic resolution ID.

**Text Preprocessing and Feature Extraction**:

- TF-IDF and n-gram (uni-, bi-, tri-gram) features are extracted to capture term importance.
- Pretrained FastText embeddings are used to obtain dense semantic representations of tickets.
- Additional linguistic features, such as part-of-speech tags and stopword filtering, are employed for improved model interpretability.

**Model Training**:

- Three base models are trained: LDA for topic clustering, a Siamese neural network for one-shot learning using triplet loss, and an Index Embedding model for joint representation learning of issue-resolution pairs.
- Each model outputs a predicted resolution class for each ticket in the test set.

**Ensemble Integration**: The outputs of the base models are fused using a logistic regression meta-learner. This stacking-based ensemble model combines strengths of individual predictors while compensating for their weaknesses.

**Validation**:

- Cross-validation and confusion matrices are used to assess model stability and detect class imbalances or label confusion.
- SME (Subject Matter Expert) validation is conducted to qualitatively assess the semantic coherence of the generated clusters.

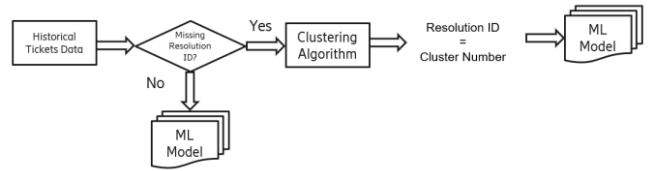

Fig. 1. Block diagram of the training phase of the proposed ML solution to identify resolutions for problem tickets

A block diagram of the training phase of the proposed approach is provided in Figure 1. It illustrates the end-to-end data processing pipeline for model training. It begins with raw incident tickets, which are preprocessed through NLP techniques such as tokenization, stemming, and FastText embedding. If resolution IDs are missing, clustering is applied to create synthetic labels. Multiple models, including LDA, Siamese network, and Index Embedding, following [1], are then trained. Their outputs are combined through an ensemble model to produce a final prediction.

### B. Inference phase

In the deployment stage, the trained ensemble model is used to predict resolutions for newly arriving tickets. The inference pipeline includes:

**Preprocessing**: Incoming tickets undergo the same preprocessing steps as the training data to ensure feature consistency.

**Resolution Prediction**:

- The ensemble model assigns the most probable resolution cluster or ID to the ticket.
- If the summed prediction probability across the top-N predictions is below a threshold

(empirically set at 30%), the ticket is flagged as "low confidence."

**Fallback Mechanism**:

- For low-confidence predictions, the system retrieves semantically similar historical tickets from the training set using cosine similarity on FastText embeddings. Alternatively, approximate nearest neighbour search (e.g., using FAISS) can also be used.
- These similar tickets, along with their resolutions, are surfaced via the dashboard for manual review.

**Logging and Monitoring**: All predictions are logged, along with metadata such as confidence scores, time of prediction, and ticket attributes, which are later used for retraining or audit purposes.

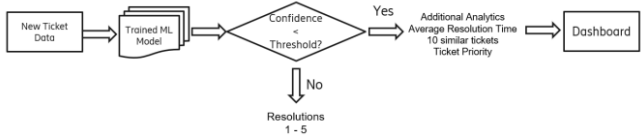

Fig. 2. Block diagram of the inference phase of the proposed ML solution to identify resolutions for problem tickets

Fig. 2 represents the workflow for the inference phase, comprising real-time prediction on incoming tickets. New tickets undergo identical preprocessing as in training. The ensemble model predicts the most suitable resolution. If the confidence is low, fallback mechanisms are triggered, which include retrieving semantically similar tickets using cosine similarity or ANN-based methods. Results and metadata are logged and made available via the PowerBI dashboard.

*C. Dashboard Integration*

A PowerBI-based dashboard is tightly integrated with the prediction pipeline to provide real-time visibility and analytical insight into system performance. The key functionalities of the dashboard include:

**Cluster Quality Analysis**:

- Displays representative resolutions from each cluster and the ticket frequency distribution across clusters.
- Highlights clusters with high variance or sparsity for further review.

**Prediction Diagnostics**:

- Visualizes prediction confidence histograms.
- Tracks model accuracy over time and surfaces common misclassifications.
- Tracks the prediction accuracy performance across resolution IDs.

**Similarity Explorer**: For low-confidence predictions, presents a list of most similar past tickets along with their resolutions to guide manual triaging.

**Operational KPIs**: Time to resolution, ticket volumes per category, anomaly trends, and data drift alerts.

This helps service desks interpret results, detect data drift, and optimize workflows.

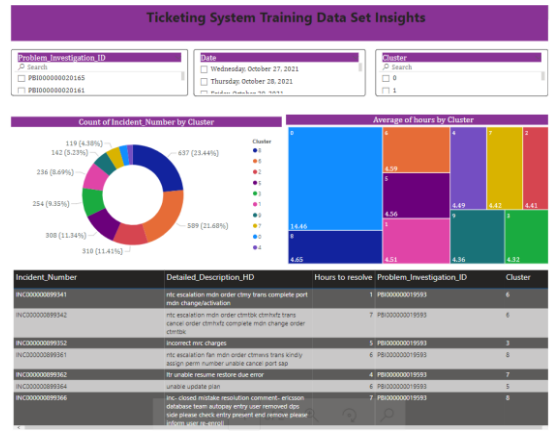

Fig. 3. PowerBI Dashboard to display the issues in the ticketing system in real time

Fig. 3 shows a screenshot of an integrated PowerBI dashboard. It displays real-time analytics such as cluster assignments, ticket volumes, average resolution time, prediction confidence, and similar ticket recommendations. This tool supports both operational monitoring and business intelligence.

Outcomes of the powerBI dashboard include the following:

- To check sparsity of cluster and identify uniqueness of cluster
- To understand the probability of a ticket falling in the cluster
- To show the TIPS algorithm performance with respect to each resolution ID
- To observe the time taken to log an incident and resolve the incident

*D. High Availability Deployment Architecture*

To ensure production-grade robustness and zero-downtime operation, the solution is deployed using a cloud-native, containerized infrastructure, including the following:

**Orchestration with Kubernetes**:

- Kubernetes manages pods for model serving, retraining pipelines, and APIs.
- Horizontal and vertical auto-scaling are enabled based on traffic and resource usage.

**Reproducible Deployments with Helm**: Helm charts encapsulate deployment configurations, simplifying environment provisioning, updates, and rollback.

**Workflow Automation with Argo**:

- Argo Workflows automate the periodic retraining pipeline, including preprocessing, model retraining, evaluation, and redeployment.
- Retraining frequency is adjustable (e.g., weekly, monthly) and is triggered based on data drift metrics.

**Experiment Tracking with MLFlow**:

- MLFlow tracks experiment parameters, metrics, and model versions.

- Comparison dashboards help select the best performing models for deployment.

**Persistent Storage and State Management**: Kubernetes Persistent Volumes ensure resilience against pod failures and data loss.

This architecture ensures a scalable, self-healing, and observable ML system, capable of processing and predicting resolutions for incoming tickets in near real-time.

This architecture allows us to ensure:

- Reliability: Services continue uninterrupted even in the face of node failures.
- Scalability: New model versions or microservices can be rolled out with zero downtime.
- Automation: Retraining pipelines are scheduled, monitored, and auto-triggered using Argo.
- Observability: MLFlow dashboards and logging pipelines provide insight into model performance, retraining status, and service health.

This infrastructure has been validated in production and supports real-time resolution prediction at scale.

We ensure system reliability and scalability in the following way:

- Kubernetes manages application containers.
- Helm charts automate deployment configurations.
- Persistent storage solutions guarantee data availability.
- Argo workflows facilitate automated retraining pipelines, monitored via MLFlow.

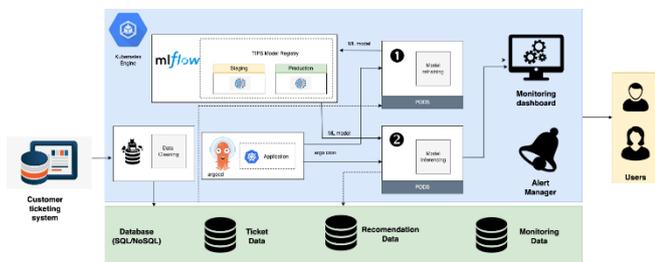

Fig. 4. Architecture of the solution with high availability

Fig. 4 shows the architecture of the solution with high availability. It depicts the infrastructure used to ensure scalability and fault tolerance. Core components include Kubernetes pods managing model serving, retraining pipelines, and API endpoints. Persistent volumes safeguard data, and Helm charts automate deployment. Load balancing and auto-scaling maintain system responsiveness under variable loads.

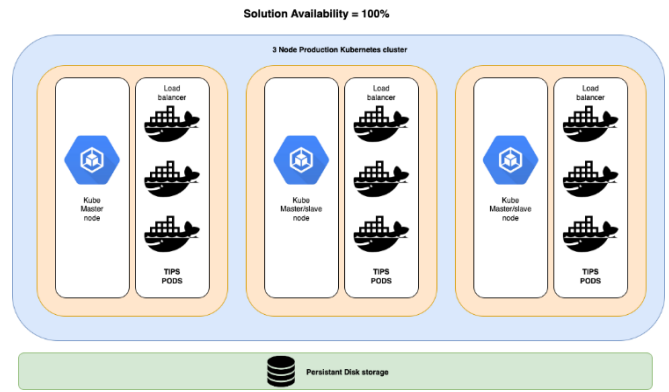

Fig. 5. Architecture of the solution for high availability showing the kubernetes notes and persistent storage

Fig. 5 provides a more granular view of the system architecture. It highlights how Kubernetes nodes interact with persistent volumes, ensuring no data loss during node failure. The architecture supports blue-green deployment strategies, enabling seamless updates and minimal downtime.

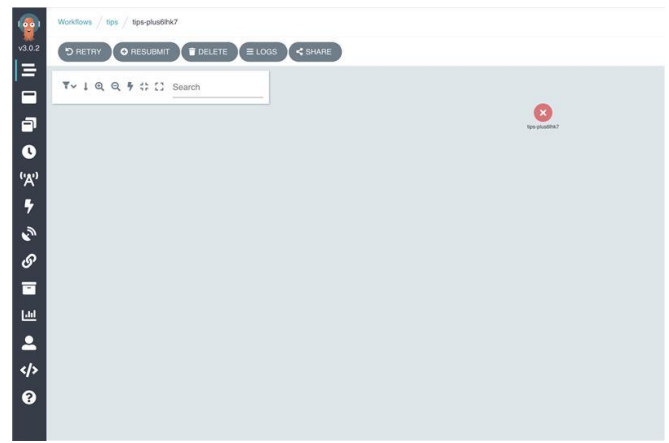

Fig. 6. Screenshot of an Argo workflow

Fig. 6 shows an actual execution trace of an Argo workflow used for periodic retraining. Each step, from data preprocessing to model evaluation, is represented as a node. The visual trace allows developers to identify bottlenecks or failures in the pipeline quickly.

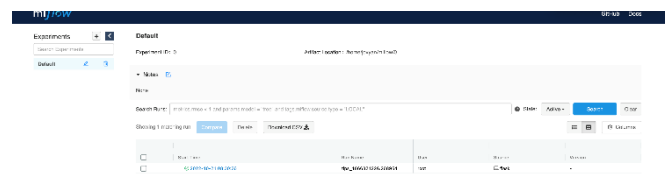

Fig. 7. Screenshot of MLFlow for retraining the model

Fig. 7 shows a screenshot of an MLFlow pipeline. Here, MLFlow is used to track multiple training experiments. Key metrics such as accuracy, loss, and parameter configurations are displayed, allowing side-by-side comparisons. This enables data scientists to make informed decisions about which model version to promote to production.

## V. RESULTS

We evaluated the proposed resolution recommendation framework on two datasets: the public Bitext customer support dataset [2] and a proprietary dataset from a major telecom operator. The telecom dataset initially contained 7,533 records, which were reduced to 4,459 after preprocessing steps such as removing null values and incomplete tickets.

To generate resolution classes in the absence of structured resolution IDs, we applied clustering algorithms including K-Means and Gaussian Mixture Models (GMM) on the resolution descriptions. The clustering process effectively grouped semantically similar resolutions and created proxy resolution IDs, enabling the subsequent supervised learning steps.

Supervised classification models, specifically XGBoost and LightGBM, were then trained on an 80:20 train-test split of the data, using 10 resolution clusters as labels. These models achieved a high precision of 98% and and recall of 98%, demonstrating the robustness of the learned representations and confirming the predictive quality of the features extracted from ticket descriptions.

The performance of the ensemble model, which combined predictions from LDA, Siamese Network, and Index Embedding models using a logistic regression meta-learner, was also manually verified through Subject Matter Expert (SME) validation. The clusters were found to be coherent, with minimal semantic overlap between resolution groups.

Further, the PowerBI dashboard provided valuable operational insights, including cluster sparsity, resolution ID performance, and prediction confidence for new incoming tickets. The dashboard also served as a fallback tool, presenting semantically similar historical tickets for manual review in cases where prediction confidence was low (below a 30% threshold).

From a systems perspective, the solution was successfully deployed in a production-like Kubernetes environment. High availability and automated retraining were achieved through container orchestration using Kubernetes, deployment scripting via Helm charts, and workflow automation with Argo. Model tracking and experiment management were handled through MLFlow, ensuring traceability and reproducibility across deployments.

This end-to-end deployment pipeline enables near real-time resolution recommendations, while also allowing for periodic retraining and feedback loop integration. The system is scalable and modular, capable of supporting enterprise-grade incident management systems.

## VI. ALTERNATIVE AND EMERGING APPROACHES

While the proposed solution presents a robust and scalable framework for incident resolution recommendation, recent developments in machine learning, deep learning, and statistical modelling provide additional avenues for addressing this task. These methods can either complement or serve as alternatives to the current pipeline, particularly in environments where data constraints, user interaction needs, or domain adaptation are critical.

### A. Large Language Models (LLMs)

Large Language Models (LLMs) such as GPT-4, LLaMA, and Mistral represent a transformative shift in natural language processing. They can be employed in several capacities:

- Retrieval-Augmented Generation (RAG): By pairing an LLM with a vector-based document retriever (e.g., Faiss or ElasticSearch), the system can surface relevant historical tickets and generate candidate resolutions with contextual understanding.

- Few-Shot Prompting and In-Context Learning: Pretrained LLMs can infer resolutions using a handful of example ticket-resolution pairs in the prompt, without requiring task-specific fine-tuning.

- Generative Chatbots: LLMs can support conversational resolution systems, enabling service agents or customers to interactively diagnose and resolve issues through dialogue.

- Domain Fine-Tuning: Using parameter-efficient tuning strategies such as LoRA or adapters, LLMs can be adapted to telecom-specific vocabulary and style with minimal resources.

LLMs reduce dependence on structured fields and provide robustness to paraphrased or semantically similar inputs. However, they require careful consideration of latency, interpretability, and data privacy in enterprise settings.

### B. Transformer-Based Sentence Encoders

Sentence-level transformers such as Sentence-BERT, DistilRoBERTa, and MiniLM can be used to encode both ticket descriptions and resolutions into fixed-size embeddings. These embeddings can power semantic search, clustering, or supervised classification pipelines. Compared to traditional TF-IDF or word embeddings, transformer-based encoders offer significantly improved performance on semantic similarity tasks.

### C. Hierarchical Attention Networks (HAN)

In cases where tickets include multi-sentence descriptions or chat transcripts, Hierarchical Attention Networks can be leveraged to model word-level and sentence-level dependencies. These architectures are particularly effective in focusing on the most informative portions of text, which is useful when noisy or verbose ticket narratives are common.

### D. Graph Neural Networks (GNNs)

When metadata such as customer IDs, time stamps, or linked ticket chains is available, the incident database can be represented as a graph. GNNs like Graph Convolutional Networks (GCNs) can model relationships between tickets, enabling information propagation across related incidents and improving resolution prediction accuracy for sparse cases.

### E. Sequence-to-Sequence (Seq2Seq) Models

Seq2Seq models with attention mechanisms (e.g., LSTM with attention, or transformer-based encoder-decoders) can be applied to generate resolution narratives given a ticket description. These models offer fine-grained control over the style and specificity of generated text and may be used to complement retrieval-based methods in generative pipelines.

*F. Advanced Topic Models*

Beyond Latent Dirichlet Allocation (LDA), more expressive topic models such as Correlated Topic Models (CTM) or Hierarchical Dirichlet Processes (HDP) can be used when the number or structure of incident types is not known a priori. These models capture topic correlations and support non-parametric clustering, offering greater flexibility than fixed-topic models.

*G. Probabilistic Soft Logic (PSL)*

PSL enables reasoning over soft constraints and domain knowledge, using logic-based templates combined with probabilistic inference. For instance, rules such as "if a ticket mentions 'payment delay' and 'invoice' it is likely related to 'billing system'" can be encoded and learned from weakly labelled data. PSL models are interpretable and suitable for integrating human expertise into the pipeline.

*H. Meta-Learning Techniques*

To handle rapid changes in incident types across service desks or customers, meta-learning approaches such as Model-Agnostic Meta-Learning (MAML) or Prototypical Networks can be used. These models allow quick adaptation to new tasks with limited data, which is useful when onboarding new clients or supporting custom ticket taxonomies.

*I. Variational and Denoising Autoencoders*

Autoencoders can be used to learn latent representations of both ticket and resolution texts. By measuring the reconstruction loss or latent-space distance between new and historical tickets, the system can flag anomalous or previously unseen issues. This serves as an auxiliary mechanism to detect data drift and model failure cases.

*J. Multimodal Learning Models*

When additional modalities such as screenshots, log files, or error codes are available, multimodal architectures can be designed to jointly learn from text and structured data. For example, CNNs can process images while transformers handle text inputs, with a shared embedding space enabling resolution prediction. This approach is particularly valuable in IT and field-support ticketing systems.

*K. Summary*

Incorporating these advanced methods can substantially enhance the resolution recommendation pipeline in terms of accuracy, adaptability, and user interaction. Hybrid strategies, such as using LLMs for fallback resolution generation, or combining transformer encoders with statistical models—are particularly promising. Future iterations of the current system may benefit from progressive integration of these models, subject to empirical validation and enterprise deployment constraints.

VII. CONCLUSION AND FUTURE WORK

We have presented a complete machine learning pipeline to predict resolutions for problem tickets, accounting for data inconsistencies, missing fields, and operational constraints. The system is accurate, scalable, and production-ready.

Future enhancements will include the following:

- Incorporating transformer-based models for improved generalization.
- Deploying LLM-powered chatbots for interactive support.
- Designing smarter fallback strategies based on semantic similarity.
- Enriching the dashboard with predictive analytics and anomaly detection.

REFERENCES

[1] Ferland N, Sun W, Fan X, Yu L, Yang J. Automatically resolve trouble tickets with hybrid NLP. In: 2020 IEEE Symposium Series on Computational Intelligence (SSCI); 2020 Dec 1–4; Canberra, Australia. IEEE; 2020. p. 1334–41.

[2] Huggingface. Bitext Customer Support LLM Training Dataset. [Online]. Available: https://huggingface.co/datasets/bitext/Bitext-customer-support-llm-chatbot-training-dataset

[3] Marzo N. Natural Language Processing Model for Log Analysis to Retrieve Solutions for Troubleshooting Processes [Master's thesis]. Stockholm: KTH Royal Institute of Technology; 2021.

[4] Rocha VG da, Kulkarni A, Rocha AAA da. Classifying customer complaints of a large fixed broadband service provider using machine learning. In: 2023 International Wireless Communications and Mobile Computing Conference (IWCMC); 2023 Jun 19–23; Marrakesh, Morocco. IEEE; 2023. p. 1334–9.

[5] Almarzooqi M. Automated Prioritization and Routing of IT Support Tickets Using Machine Learning [Master's thesis]. Rochester (NY): Rochester Institute of Technology; 2025.

[6] Feng L, Senapati J, Liu B. TaDaa: real time Ticket Assignment Deep learning Auto Advisor for customer support, help desk, and issue ticketing systems [Internet]. arXiv preprint arXiv:2207.11187v2; 2022 Jul 22 [cited 2025 Jul 25]. Available from: https://arxiv.org/abs/2207.11187